# Distance Measurement for UAVs in Deep Hazardous Tunnels

Vishal Choudhary, Shashi Kant Gupta, Shaohui Foong, Hock Beng Lim
Centre for Smart Systems
Singapore University of Technology and Design
[vishal_choudhary | foongshaohui | limhb]@sutd.edu.sg

*Abstract* — The localization of Unmanned Arial Vehicles (UAVs) in deep tunnels is extremely challenging due to their inaccessibility and hazardous environment. Conventional outdoor localization techniques (such as using GPS) and indoor localization techniques (such as those based on WiFi, Infrared (IR), Ultra-Wideband, etc.) does not work in deep tunnels. We are developing a UAV-based system for the inspection of defects in the Deep Tunnel Sewerage System (DTSS) in Singapore. To enable the UAV localization in the DTSS, we have developed a distance measurement module based on the optical flow technique. However, the standard optical flow technique does not work well in tunnels with poor lighting and lack of features. Thus, we have developed an enhanced optical flow algorithm with prediction, to improve the distance measurement for UAVs in deep hazardous tunnels.

## I. Introduction

In recent years, Unmanned Aerial Vehicles (UAVs) are increasingly used for many commercial and defence applications. For outdoor applications, UAVs usually rely on GPS for localization. In GPS denied environments, the UAVs' navigation system will not work well due to degraded GPS or no GPS signal, and other techniques are required for localization [1]. We are currently developing a UAV-based system for the inspection of defects in the Deep Tunnel Sewerage System (DTSS) in Singapore. The DTSS is a good example of such GPS denied environments.

Optical flow sensors have been used for distance measurements through the calculation of velocities [2]. However, it is essential for optical flow sensors to operate under good lighting in order to provide accurate measurements. The lighting issue can be solved by using an onboard lighting mechanism such as strong LEDs. Power consumption and distance from the ground are two factors that affect optical flow sensor performance.

A smart optical flow sensor, Px4Flow [3], has been developed using low-cost components. The Px4Flow is available as a commercial product. It computes optical flow and compensates for rotations. The flow estimation performance of Px4Flow matches typical optical flow sensors and it works well in environments with good lighting and textures. But when the illumination quality is bad, Px4flow does not perform well and gives zero flow values. In this work, we developed an algorithm to predict these missing or zero flow values using some previous acceleration and velocities.

## II. Algorithm

We are using a Px4flow optical flow sensor which sum up the accumulated pixel flow values between two I2C readouts. The sensor is mounted on our UAV platform for operation in the DTSS. The following mathematical formula is used to calculate the displacement:

$$s = \theta \times r \quad \begin{array}{l} S - Displacement \\ \theta - Angular\ Displacement \\ r - Radius \end{array}$$

In the above equation, the accumulated pixel flow in radians from Px4Flow is 'θ' and distance measured by the Lidar Lite is 'r'. The PX4Flow comes with a sonar sensor for distance measurement to the surface, which is not good enough in terms of accuracy. Instead, we use Lidar Lite which measures the distance to the surface more accurately. We also use a scalar equation for better distance estimation.

$$r = 1.07 \times r_{raw} - 100$$
$$r_{raw} - raw\ distance\ data\ from\ Lidar$$

The Px4Flow sensor is noisy and susceptible to low light intensities and interferences. We use the Cree XHP-50 LEDs to provide higher light intensities in a dark tunnel. Sometimes if the tunnel surface is low in features, the Px4Flow might not able get good quality flow values, which leads to inaccurate displacement. To overcome this limitation, we developed a prediction algorithm that uses the continuously calculated accelerations data from the past when the optical quality is good. Accelerations are calculated using the following formula:

$$a = \frac{dv}{dt}$$

When the quality >100, we use the velocity received directly from Px4Flow sensor, whereas when quality <100, we use the velocity predicted using the approximated acceleration. In our algorithm, we record the accelerations and velocities data for some specified number when the quality is good (Q >100). We record the last *j* velocities when the quality is good. We let the past velocities to be stored in an array of **v[j]**, acceleration to be in **a[j]**, time recorded at respective velocity to be in **t[j]**, and velocity predicted to be in **v_predicted[j]**. We formed a linear model out of this **v_predicted** velocities to calculate the distance:

$$a[i] = \frac{(v[j] - v[i])}{t[j] - t[i]} \ for\ i < j$$

$$a[j] = 0$$

$$v_{predicted}[i] = v[i] + (t[p] - t[i]) \times a[i]$$

$$where\ t[p] - Time\ of\ Prediction$$

## III. Implementation

Figure 1 shows the system architecture of our distance measurement module. We have a 3D printed detachable module which can be attached to the UAV. The module consists of an optical flow sensor Px4Flow [3], Lidar Lite V3, Two CREE LEDs, a Teensy microcontroller, and a LiPo battery. The fully assembled module is shown in Figure 2.

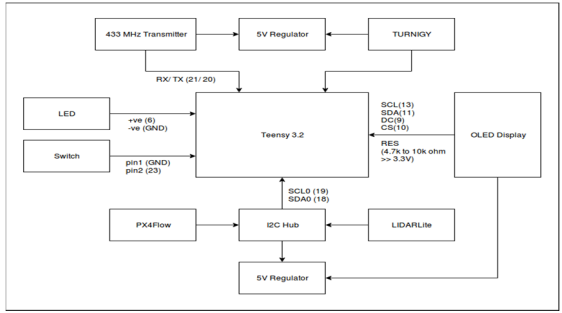

Figure 1: System Architecture

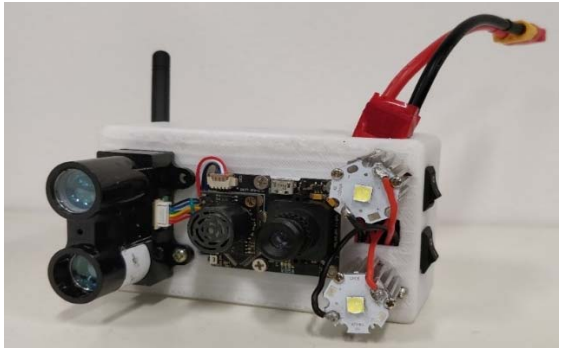

Figure 2: Distance Measurement Module

## IV. RESULTS

| | Algo/Lighting option | Ground truth | Without LEDs | With LEDs only | With Structured light only | With LEDs and Structured light |
|---|---|---|---|---|---|---|
| **Side wall** | Standard Optical Flow | 50 | 0.0 | 0.0 | 0.0 | 2.4467 |
| | With Prediction | 50 | 0.0 | 0.0 | 0.0 | 6.2265 |
| **Floor** | Standard Optical flow | 50 | 20.07 | 46.48 | 20.55 | 47.45 |
| | With Prediction | 50 | 23.37 | 50.42 | 21.609 | 50.946 |
| **Ceilings** | Standard Optical flow | 50 | 32.83 | 13.03 | 30.91 | 17.15 |
| | With Prediction | 50 | 35.35 | 18.14 | 33.50 | 20.73 |

**Table 1: Distance measurement results in passenger tunnel**

We have tested our module in a passenger tunnel in Singapore which has poor lighting and surface texture conditions. We evaluated our module by pointing it towards different surfaces, i.e. the tunnel's sidewall, ceiling and floor. During this experiment, we moved our module for 50 meters pointing towards the three aforementioned directions of the passenger tunnel and compare our algorithm results with that measured by the standard optical flow. We repeated all the tests with illumination, without illumination and structural light. From the results in Table 1, it is clear that the performance of the module with high illumination is good but degrades with poor illumination.

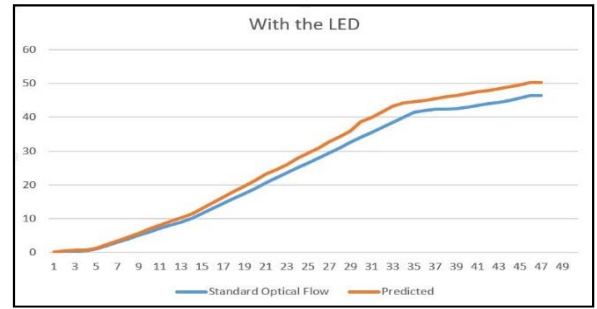

**Figure 3: Displacement on Tunnel's floor with LED lightning**

The sidewall of the tunnel has reflective white tiles, which makes it lacking in features. Due to the lacking of features, our distance measurement module is not able to get good flow values. In this case, we have tried to use a laser-based structure light to put some pattern on the wall. We found that the optical flow module is able to measure some pixel flow but that is also not enough for accurate distance measurement. We can see from Table 1 that both algorithms are giving zero displacements with and without illumination. After using structural light, the module is able to measure some displacement but that is negligible. In the second run, we pointed our module towards the ceiling of the passenger tunnel, which is also white but with some texture. We repeated the experiments and found that without using LED, the module gives better estimation of displacement but not accurate. If we turn on the LED, The displacement is reduced because the features of the surface vanished due to the high brightness of the LED. In the third run, we carried out the same experiment by pointing the module towards the floor. The tunnel's floor has some texture which makes suitable for our module measurement. Using standard displacement algorithm, it shows the displacement of 46.48meters for the same environmental parameters, while our algorithm provides a more accurate measurement of 50.98 meters. Figure 3 shows the distance measurement when the module is pointed towards the floor surface with illumination. In future, we will investigate the usage of IMU data to help predict the missing velocity or acceleration. We can use the accelerometer to predict missing flow values when the optical flow quality is bad.

## V. DEMO SETUP

We have developed a fully functional distance measurement module prototype as shown in Figure 2. We will demonstrate the distance measurement operation of our module. The distance measurement module will be powered using an external LiPo battery. To visualize the distance travelled, we have attached a small OLED display to our module. We will demonstrate how lighting condition affects the distance measurement.